\pdfoutput=1
\documentclass[11pt]{article}

\usepackage[]{acl}

\usepackage{times}
\usepackage{tabularx}
\usepackage{latexsym}
\usepackage{graphicx}
\usepackage{makecell}

\usepackage[T1]{fontenc}
\usepackage[utf8]{inputenc}

\usepackage{microtype}
\usepackage{amssymb}% http://ctan.org/pkg/amssymb
\usepackage{pifont}% http://ctan.org/pkg/pifont
\usepackage{booktabs}
\newcolumntype{H}{>{\setbox0=\hbox\bgroup}c<{\egroup}@{}}
\definecolor{Gray}{gray}{0.9}
\definecolor{Green}{rgb}{0.67, 0.88, 0.69}
\definecolor{darkgray}{rgb}{0.66, 0.66, 0.66}
\definecolor{lavendergray}{rgb}{0.81, 0.81, 0.77}
\newcolumntype{Y}{>{\centering\arraybackslash}X}
\title{Knowledge-Augmented Language Models for Cause-Effect Relation Classification}

\author{
\fontsize{12pt}{12pt}\selectfont
 \makecell{Pedram Hosseini$^{1}$ \quad David A. Broniatowski$^{1}$ \quad Mona Diab$^{1,2}$}\\
 \fontsize{12pt}{12pt}\selectfont
 \makecell{$^{1}$The George Washington University \quad $^{2}$Meta AI}\\ 
\fontsize{12pt}{12pt}\selectfont
\makecell{\texttt{\{phosseini,broniatowski\}@gwu.edu, mdiab@fb.com}}
}

\begin{document}
\maketitle
\begin{abstract}
Previous studies have shown the efficacy of knowledge augmentation methods in pretrained language models. However, these methods behave differently across domains and downstream tasks. In this work, we investigate the augmentation of pretrained language models with commonsense knowledge in the cause-effect relation classification and commonsense causal reasoning tasks. After automatically verbalizing ATOMIC$^{20}_{20}$, a wide coverage commonsense reasoning knowledge graph, and GLUCOSE, a dataset of implicit commonsense causal knowledge, we continually pretrain BERT and RoBERTa with the verbalized data. Then we evaluate the resulting models on cause-effect pair classification and answering commonsense causal reasoning questions. Our results show that continually pretrained language models augmented with commonsense knowledge outperform our baselines on two commonsense causal reasoning benchmarks, COPA and BCOPA-CE, and the Temporal and Causal Reasoning (TCR) dataset, without additional improvement in model architecture or using quality-enhanced data for fine-tuning.
\end{abstract}

\section{Introduction}
\label{sect:introduction}
Automatic extraction and classification of causal relations in the text have been important yet challenging tasks in natural language understanding. Early methods in the 80s and 90s~\cite{joskowicz1989deep,kaplan1991knowledge,garcia1997coatis,khoo1998automatic} mainly relied on defining hand-crafted rules to find cause-effect relations. Starting 2000, machine learning tools were utilized in building causal relation extraction models~\cite{girju2003automatic,chang2004causal,chang2006incremental,blanco2008causal,do2011minimally,hashimoto2012excitatory,hidey-mckeown-2016-identifying}. Word-embeddings and Pretrained Language Models (PLMs) have also been leveraged in training models for understanding causality in language in recent years~\cite{dunietz2018deepcx,pennington2014glove,dasgupta2018automatic,gao2019modeling}. Knowledge Graphs (KGs) have been also used in combination with pretrained language models to address commonsense reasoning~\cite{li2020guided,guan2020knowledge}. Despite all these efforts, investigating the true capability of pretrained language models in understanding causality in text is still an open question.

\begin{figure}[t]
\centering
\includegraphics[scale=0.72]{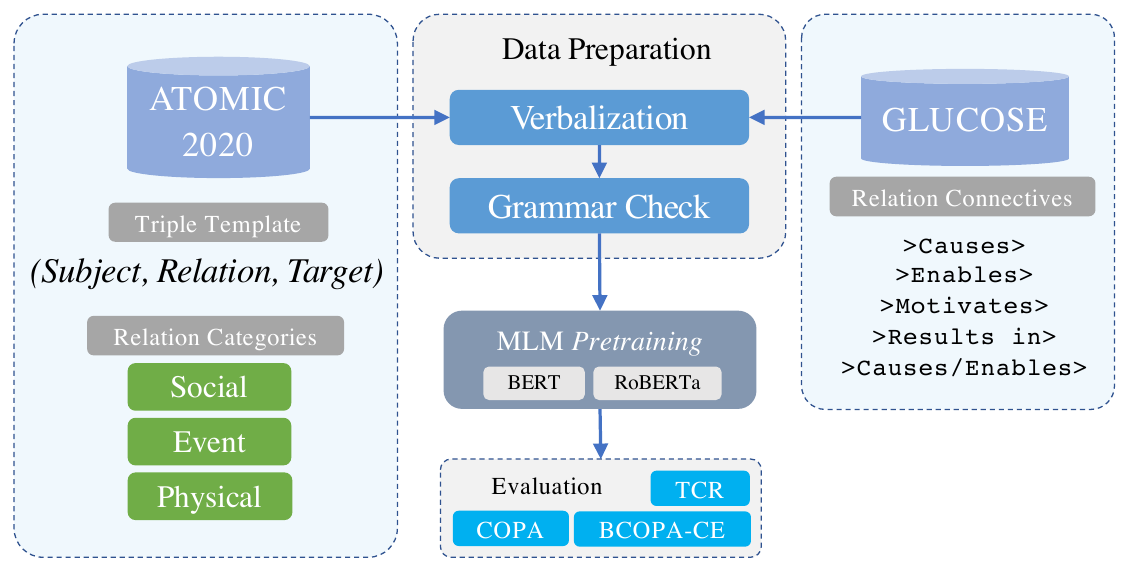}
\caption{\label{fig:method}Overview of our proposed framework to continually pretrain PLMs with commonsense knowledge.}
\end{figure}

In this work, motivated by the success of continual pretraining of PLMs for downstream tasks~\cite{gururangan2020don}, we explore the impact of commonsense knowledge injection as a form of continual pretraining for causal reasoning and \textit{cause-effect} relation classification.
% We hypothesize that continual pretraining of PLMs using commonsense knowledge will improve performance on Cause-Effect relation classification. 
%Moreover, models with a significantly fewer number of parameters (BERT) compared to large PLMs such as DeBERTa~\cite{he2020deberta}, Google T5~\cite{raffel2019exploring}, or GPT-3~\cite{brown2020language} can benefit from such a continual pretraining.
It is worth highlighting that even though there are studies to show the efficacy of knowledge injection with continual pretraining for commonsense reasoning~\cite{guan2020knowledge}, performance of these techniques is very dependent on the domain and downstream tasks~\cite{gururangan2020don}. And, to the best of our knowledge, there are limited studies on the effect of commonsense knowledge injection on \textit{causal} relation classification~\cite{dalal2021enhancing}. Our contributions are as follows:
\begin{itemize}
\itemsep0em
\item We study the performance of PLMs augmented with commonsense knowledge in the less investigated task of cause-effect relation classification.
\item We demonstrate that a simple masked language modeling framework using automatically verbalized commonsense knowledge, without any further model improvement (e.g., new architecture or loss function) or quality enhanced data for fine-tuning, can significantly boost the performance of PLMs in cause-effect pair classification.
\item We publicly release our knowledge graph verbalization codes and continually pretrained models. 
\end{itemize}
%Moreover, choosing proper data for continual pretraining in itself is very important and not the same across downstream tasks.

% In this paper, we first describe our method (\S\ref{sec:method}) and delineate the process of converting knowledge graph data to natural language text (\S\ref{subsec:kg-to-text}) for continual pretraining (\S\ref{subsec:pretraining}). Then, we explain our experiments (\S\ref{sec:experiments}) and present and discuss the results and next steps (\S\ref{sec:result}).

\section{Method}
\label{sec:method}
The overview of our method is shown in Figure~\ref{fig:method}.\footnote{Codes and models are publicly available at \url{https://github.com/phosseini/causal-reasoning}.} In our framework, we start by verbalizing ATOMIC$^{20}_{20}$~\cite{Hwang2021COMETATOMIC2O} knowledge graph and GLUCOSE~\cite{mostafazadeh2020glucose} to natural language texts. Then we continually pretrain BERT~\cite{devlin2018bert} and RoBERTa~\cite{liu2019roberta} using Masked Language Modeling (MLM) and evaluate performance of the resulting models on different benchmarks. We delineate each of these steps in the following sections.

%\begin{figure}[h]
%\centering
%\includegraphics[scale=0.4]{atomic_length_v1.pdf}
%\caption{\label{fig:sequence_length}Number of triples in ATOMIC2020 grouped by the number of tokens (separated by space) in verbalized triples.}
%\end{figure}

%\footnote{There are 68626, 7410, and 8473 duplicate triples in train, development, and test sets of ATOMIC$^{20}_{20}$, respectively. These duplicate triples are redundant and indicate multiple annotations for some head/relation pairs.}

\subsection{ATOMIC$^{20}_{20}$ to Text}
\begin{figure}[h]
\centering
\includegraphics[scale=0.57]{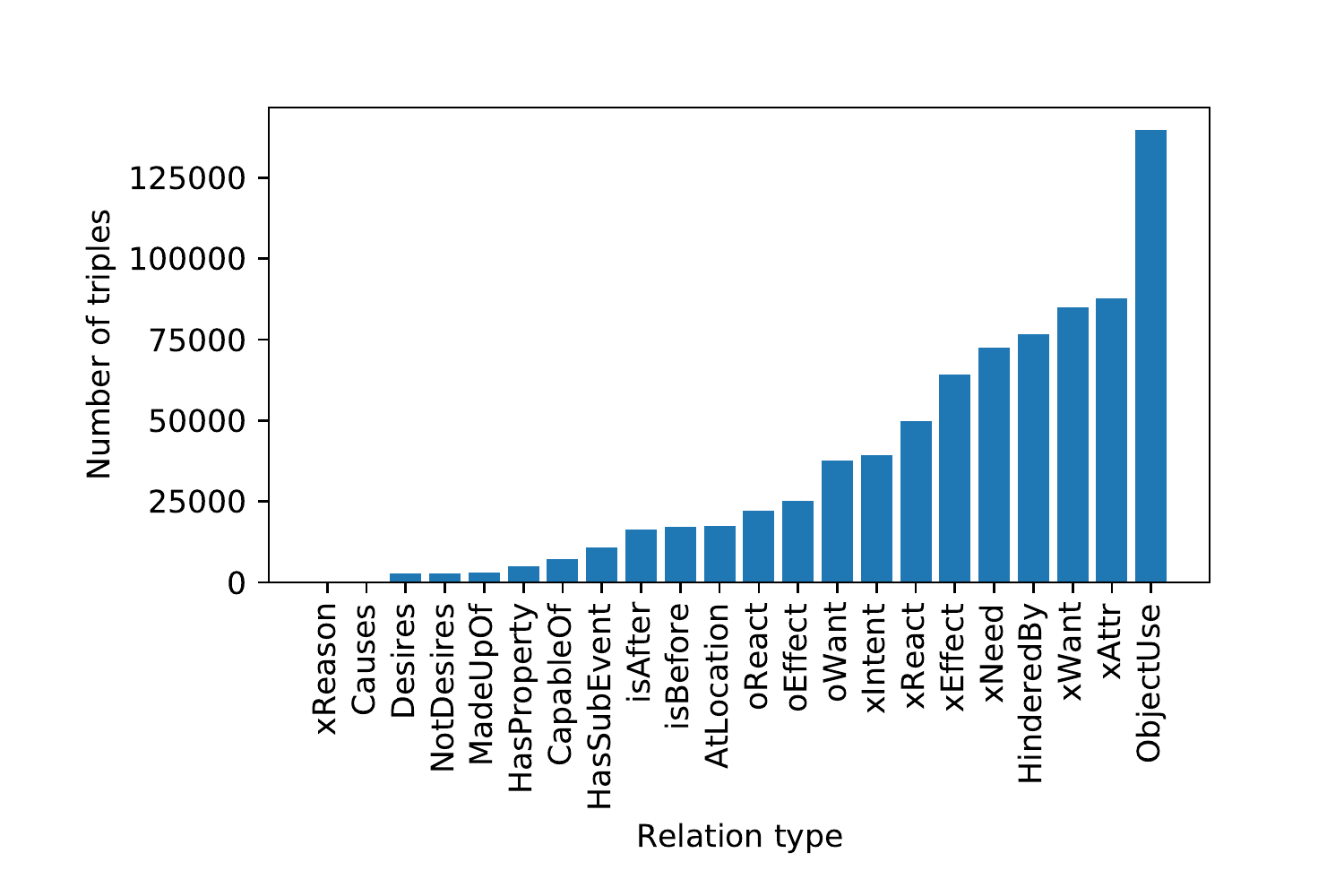} \caption{\label{fig:relations}Distribution of relation types in ATOMIC$^{20}_{20}$.}
\end{figure}

Samples in ATOMIC$^{20}_{20}$ are stored as triples in the form of \textit{(head/subject, relation, tail/target)} in three splits including train, development, and test. We only use the train and development sets here. ATOMIC$^{20}_{20}$ has 23 relation types that are classified into three categorical types including commonsense relations of social interactions, physical-entity commonsense relations, and event-centric commonsense relations. In the rest of the paper, we refer to these three categories as social, physical, and event, respectively. Distribution of these relations is shown in Figure~\ref{fig:relations}. Each relation in ATOMIC$^{20}_{20}$ is associated with a human-readable template. For example, templates for \textit{xEffect} and \textit{HasPrerequisite} are \textit{as a result, PersonX will} and \textit{to do this, one requires}, respectively. We use these templates to convert triples in ATOMIC$^{20}_{20}$ to sentences in natural language (verbalization) by concatenating the subject, relation template, and target.

Before verbalizing triples, we also remove all duplicates and ignore all triples in which the target value is \textit{none}. Moreover, we ignore all triples that include a blank. Since in masked language modeling we need to know the gold value of masked tokens, a triple that already has a blank (masked token/word) in it may not help our pretraining. For instance, in the triple: {\tt [PersonX affords another \_\_\_, xAttr, useful]} it is hard to know why or understand what it means for a person to be useful without knowing what they afforded. This preprocessing step yields in 782,848 triples with 121,681, 177,706, and 483,461 from event, physical, and social categories, respectively.

\begin{figure}[h]
\centering
\includegraphics[scale=0.52]{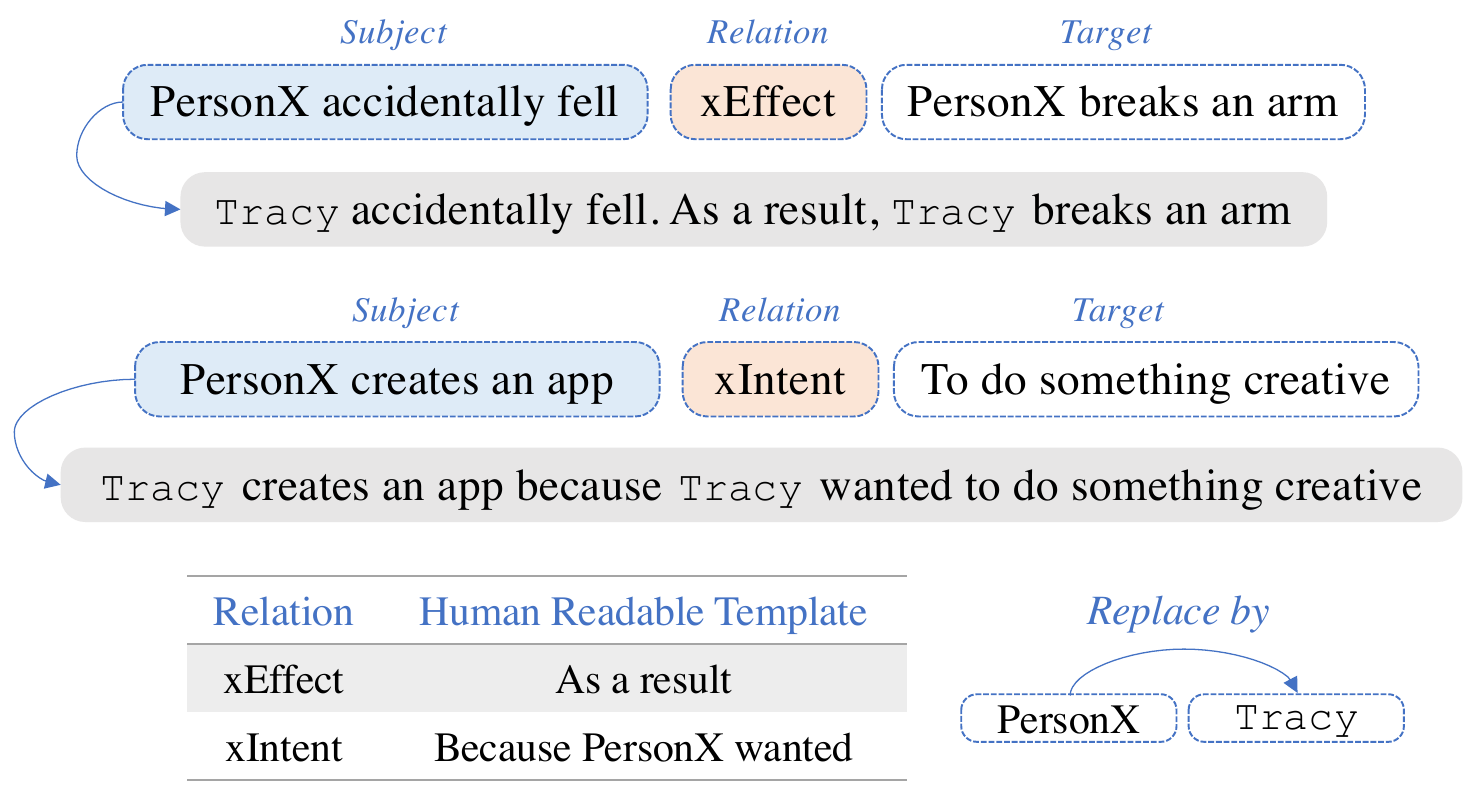}
\caption{\label{fig:atomic-conversion}Examples of converting two triples in ATOMIC$^{20}_{20}$ to natural language text (verbalization) using human readable templates. Following~\citet{sap-etal-2019-social}, we replace \textit{PersonX} with a name.}
\end{figure}

Examples of converting triples to text are shown in Figure~\ref{fig:atomic-conversion}.

\subsection{GLUCOSE to Text}
\begin{figure}[h]
\centering
\includegraphics[scale=0.65]{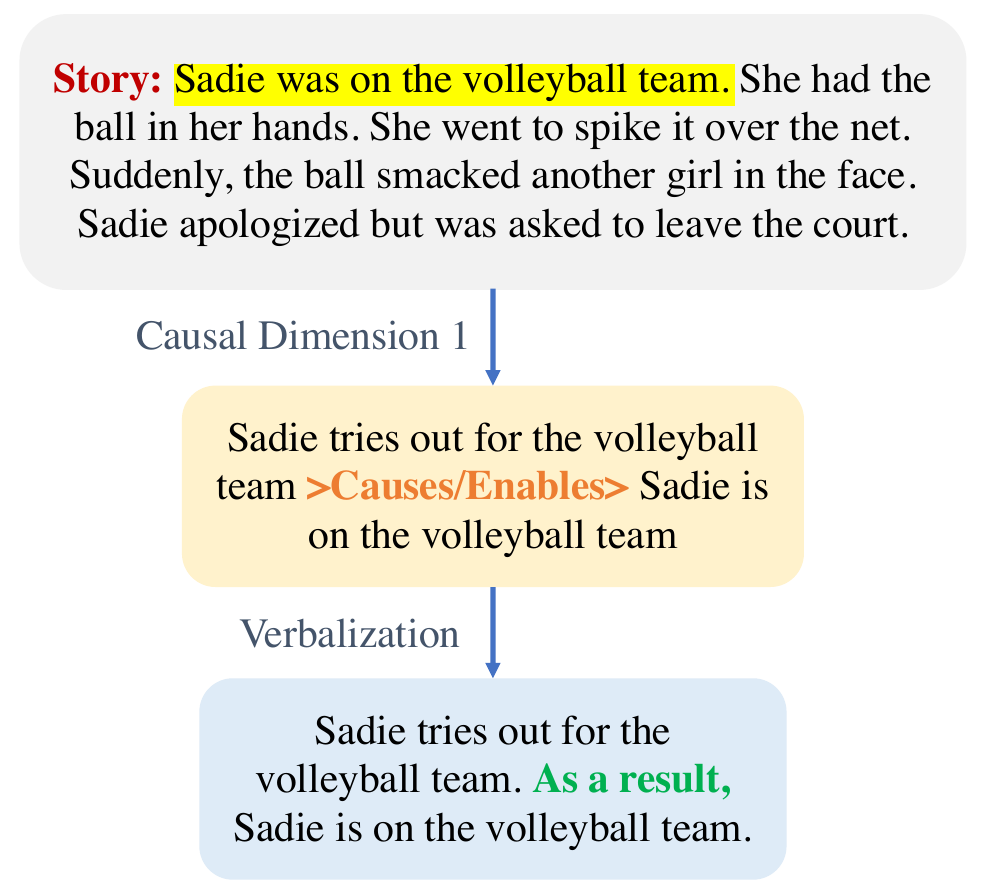}
\caption{\label{fig:glucose-conversion}Example of verbalizing GLUCOSE.}
\end{figure}

GLUCOSE is a large-scale dataset of implicit commonsense causal knowledge. Each data point in GLUCOSE includes ten dimensions of causal explanations for a selected sentence in a story with a focus on events, states, motivations, and emotions. Half of these dimensions are specific causal statements and the remaining half are general rules that capture the implicit commonsense knowledge. Using a slightly modified version of templates that are provided for causal connectives in GLUCOSE, we concatenate the two spans in a causal relation with each relation's template to form a verbalized sample. The causal connectives in GLUCOSE include: {\tt [>Causes/Enables>, >Motivates>, >Enables>, >Causes>, >Results in>]}. Verbalization of a sample in GLUCOSE is shown in Figure~\ref{fig:glucose-conversion}. In the end, we randomly split the verbalized samples into train (90\%) and development (10\%) sets. 

\subsection{Checking Grammar}
When we verbalize samples in ATOMIC$^{20}_{20}$ and GLUCOSE to natural language text, ideally we want to have grammatically correct sentences. Human readable templates provided by ATOMIC$^{20}_{20}$ and GLUCOSE are not necessarily rendered in a way to always form error-free sentences.
%For example, after concatenating relation type and target in a tuple of the knowledge graph, we may have a sentence such as: \textit{As a result, PersonX wants leave} which is grammatically incorrect since there is a \textit{to} missing after \textit{wants}.
To address this issue, we use an open-source grammar and spell checker, LanguageTool,\footnote{\url{https://tinyurl.com/yc77k3fb}} to double-check our converted triples to ensure they do not contain obvious grammatical mistakes or spelling errors. Similar approaches that include deterministic grammatical transformations were also previously used to convert KG triples to coherent sentences~\cite{davison2019commonsense}. It is worth pointing out that the Data-To-Text generation (KG verbalization) itself is a separate task and there have been efforts to address this task~\cite{agarwal2021knowledge}. We leave investigating the effects of using other Data-To-Text and grammar-checking methods as future research. %see whether they improve the quality of generated text from KG can be considered as one next step.

%The grammar checking process resulted in modifying a total of 151,783 samples (\%19 of all samples).%\footnote{We make the converted samples and conversion codes publicly available. We have also flagged all the corrected/modified samples.}

\subsection{Continual Pretraining}
\label{subsec:pretraining}
As mentioned earlier, we use MLM\footnote{We use Huggingface's \textit{BertForMaskedLM}.} to continually pretrain our PLMs, \textit{bert-large-cased} and \textit{roberta-large}. We follow the same procedure as BERT to create the input data for our pretraining (e.g., number of tokens to mask in input examples). We run the pretraining using \textit{train} and \textit{development} splits in ATOMIC$^{20}_{20}$ and GLUCOSE (separately) as our training and evaluation sets, respectively, for 10 epochs on Google Colab TPU v2 using \textit{PyTorch/XLA} package with a maximum sequence length of 30\footnote{\%99.99 of verbalized instances have 30 tokens or less.} and batch size of 128. To avoid overfitting, we use early stopping with the patience of 5 on evaluation loss. We select the best model based on the lowest evaluation loss at the end of training.

\begin{figure}[h]
\centering
\includegraphics[scale=0.51]{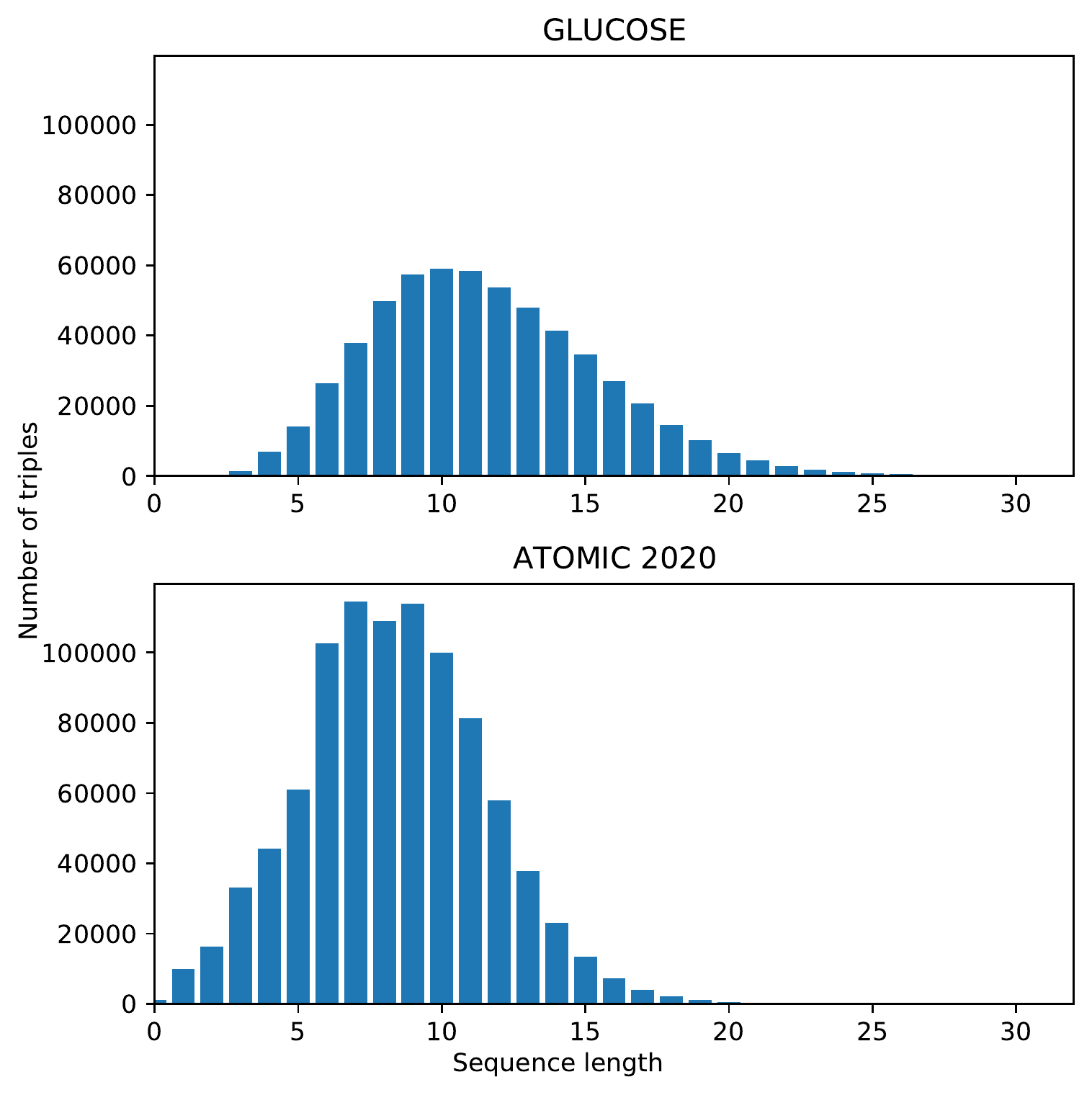}
\caption{\label{fig:glucose_atomic_sequence_length}Distribution of samples in ATOMIC$^{20}_{20}$ and GLUCOSE based on the number of tokens (separated by space).}
\end{figure}

\section{Experiments}
\label{sec:experiments}
%In our experiments, we first run a 10-fold cross-validation on the training set for tuning the hyperparameters. Then, using the best hyperparameter tuning trial, we fine-tune our models with four different random seeds using the entire training set, evaluate the fine-tuned models on the test set, and report the average performance.
%We run our experiments on two pretrained language models including 1) BERT-large-cased, and 2) CausalBERT-base. These models are different in two ways that enable us to measure the effect of our method in multiple ways. First, BERT-large model has not been continually pretrained using any additional domain-specific data while CausalBERT-base is pretrained using CausalBank which contains a large collection of causal relations. This allows us to see how adding knowledge graph data to models that are pretrained using different data have an impact on their performance on commonsense causal reasoning. Second, CausalBERT-base uses a different loss function than BERT-large and this allows us to better compare the effect of our continual pretraining with the CausalBERT-base model itself.

\subsection{Benchmarks}
\label{subsec:benchmarks}
We chose multiple benchmarks of commonsense causal reasoning and cause-effect relation classification to ensure we thoroughly test the effects of our newly trained models. These benchmarks include 1) Temporal and Causal Reasoning (TCR) dataset~\cite{ning-etal-2018-joint}, a benchmark for joint reasoning of temporal and causal relations; 2) Choice Of Plausible Alternatives (COPA)~\cite{roemmele2011choice} dataset which is a widely used and notable benchmark~\cite{rogers2021qa} for commonsense causal reasoning; And 3) BCOPA-CE~\cite{han-wang-2021-good}, a new benchmark inspired by COPA, that contains unbiased token distributions which makes it a more challenging benchmark. For COPA-related experiments, since COPA does not have a training set, we use COPA's development set for fine-tuning our models and testing them on COPA's test set (COPA-test) and BCOPA-CE. For hyperparameter tuning, we randomly split COPA's development set into train (\%90) and dev (\%10) and find the best learning rate, batch size, and number of train epochs based on the evaluation accuracy on the development set. Then using COPA's original development set and best set of hyperparameters, we fine-tune our models and evaluate them on the test set. For TCR, since there is no development set and TCR's train split is not large enough for creating train and development sets, we skip hyperparameter tuning and fine-tune all models for 10 epochs with batch size of 8 and learning rate of 2e-5 on the train set and evaluate fine-tuned models on the test set. In all experiments, we report the average performance of models across eight different random seed runs.

\subsection{Models and Baseline}
\begin{figure}[h]
\centering
\includegraphics[scale=0.71]{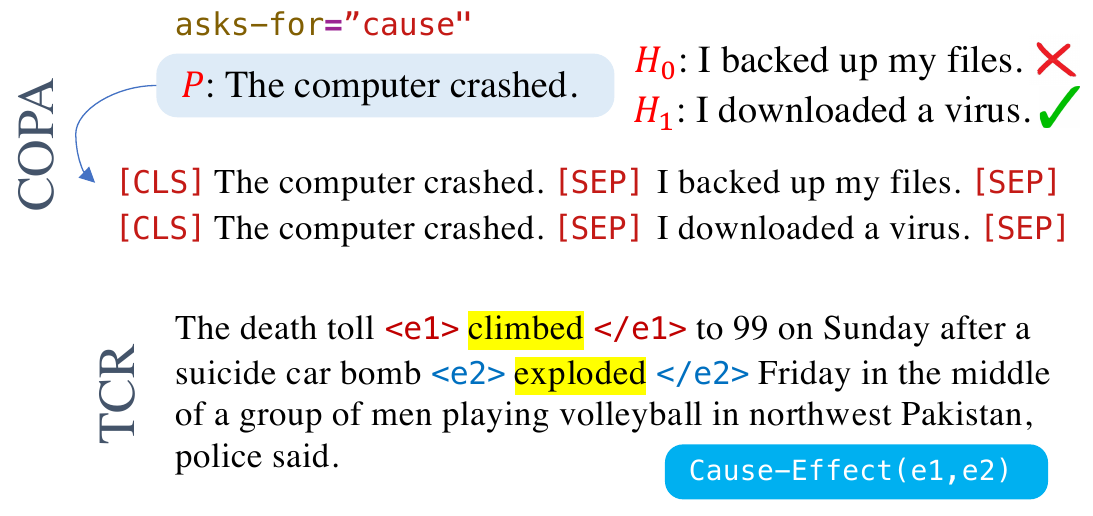}
\caption{\label{fig:copa-conversion}COPA and TCR examples. The COPA instance is converted to Multiple Choice format.}
\end{figure}

We use \textit{bert-large-cased} and \textit{roberta-large} pretrained models in our experiments as baseline. For COPA and BCOPA-CE, we convert all instances to a SWAG-formatted data~\cite{zellers2018swag} and use Huggingface's \textit{BertForMultipleChoice} --a BERT model with a multiple-choice classification head on top. And for TCR, we convert every instance by adding special tokens to input sequences as event boundaries and use the R-BERT~\footnote{We use the following implementation of R-BERT: \url{https://github.com/monologg/R-BERT}} model~\cite{wu2019enriching}. We chose R-BERT for our relation classification since it not only leverages the pretrained embeddings but also transfers information of target entities (e.g., events in a relation) through model's architecture and incorporates encodings of the target entities. Examples of COPA and TCR are shown in Figure~\ref{fig:copa-conversion}. BCOPA-CE has the same format as COPA.

\section{Results and Discussion}
\label{sec:result}
Results of our experiments on TCR are shown in Table~\ref{tab:tcr-results}. As can be seen, our best model that is continually pretrained with GLUCOSE significantly outperforms our baseline and the joint inference framework by~\citet{ning-etal-2018-joint} formulated as an integer linear programming (ILP) problem. 

\begin{table}[h]
\centering
\scalebox{0.85}{
\begin{tabular}{lcH}
\toprule
\multicolumn{1}{c}{\textbf{Model}} & \textbf{Acc (\%)}& \textbf{Best Acc (\%)} \\ \hline
Joint system~\cite{ning-etal-2018-joint} & 77.3 & - \\ \midrule \midrule
\textbf{Our Models} & & \\
BERT-Large (baseline) & 79.1$_{(0.1)}$ & 85.0 \\
ATOMIC-BERT-Large & 80.9$_{(0.11)}$ & 86.0 \\
GLUCOSE-BERT-Large & \textbf{83.9}$_{(0.02)}$ & \textbf{87.0} \\
\bottomrule
\end{tabular}
}
\caption{TCR Accuracy results.}
\label{tab:tcr-results}
\end{table}

%We initially used MLM and NSP for continual pretraining of BERT model. Based on the performance of the fine-tuned model on COPA, we observed a $\%6.7$ drop in performance compared to the vanilla BERT that is our baseline here. This drop demonstrated that the further pre-trained model with MLM and NSP not only did not learn anything new but also may have probably forgotten some of the knowledge it already had encoded. By looking at some of the examples from our pretraining data, we hypothesized such a drop in performance may be due to having NSP as one of the objectives for further pretraining. And, NSP may not be a good choice as an objective to further pretrain BERT to augment it with commonsense knowledge. As an instance, in \textit{PersonX feels like a failure. As a result, PersonX feels sad,} in the first round that we treat \textit{PersonX feels like a failure} and \textit{As a result, PersonX feels sad} as two consecutive sentences, the re-ordered sentences, \textit{As a result, PersonX feels sad. PersonX feels like a failure.} can still be viewed as consecutive since feeling sad and feeling like a failure can both be the result/cause of one another.
Results of experiments on COPA-test are shown in Table~\ref{tab:copa-results}. As can be seen, all our models significantly outperform our baselines and the performance gap between the baseline and the best model is larger for \textit{roberta} models. Also, GLUCOSE models, despite being trained with significantly fewer training data points ($\sim$70k), achieved performance on par with and even slightly better than models trained with ATOMIC$^{20}_{20}$ ($\sim$121k for event only and $\sim$780k for all three types). We also observe that continually pretrained ATOMIC$^{20}_{20}$ models using only event relations achieve almost the same performance as models trained with all three types of relations with $\sim$6X more training data points. By taking a closer look at each relation type, we realize that one reason may be the fact that event-centric relations in ATOMIC$^{20}_{20}$ specifically contain commonsense knowledge about event interaction for understating likely causal relations between events in the world~\cite{Hwang2021COMETATOMIC2O}. In addition, event relations have a relatively longer context (\# of tokens) than the average of all three relation types combined which means more context for a model to learn from.
\begin{table}[h]
\centering
\scalebox{0.9}{
\begin{tabular}{lcH}
\toprule
\multicolumn{1}{c}{\textbf{Model}} & \textbf{Acc (\%)} & \textbf{Max Acc (\%)} \\ \hline
PMI~\cite{roemmele2011choice} & 58.8 & - \\
b-l-\textit{reg}~\cite{han-wang-2021-good} & 71.1 & - \\
Google T5-base~\cite{raffel2019exploring} & 71.2 & - \\
BERT-Large~\cite{kavumba2019choosing} & 76.5 & - \\
CausalBERT~\cite{li2020guided} & 78.6 & - \\
BERT-SocialIQA~\cite{sap-etal-2019-social}$^{*}$ & 80.1 & 83.4 \\
Google T5-11B~\cite{raffel2019exploring} & 94.8 & - \\
DeBERTa-1.5B~\cite{he2020deberta} & 96.8 & - \\ \midrule \midrule
\textbf{Our Models} &  & \\
BERT-Large (baseline) & 75.5$_{(0.07)}$ & 81.6 \\
ATOMIC-BERT-Large &  & \\
\hspace{10mm}\small{{- Event, Physical, Social}} & 79.1$_{(0.03)}$ & 81.8  \\
\hspace{10mm}\small{{- Event only}} & 79.1$_{(0.01)}$ & 80.6 \\
GLUCOSE-BERT-Large & \textbf{79.9}$_{(0.02)}$ & 81.8 \\\hline
RoBERTa-Large (baseline) & 74.1$_{(0.11)}$ & 0.882 \\
ATOMIC-RoBERTa-Large &  \\
\hspace{10mm}\small{{- Event, Physical, Social}} & 83.9$_{(0.02)}$ & 85.6 \\
\hspace{10mm}\small{{- Event only}} & 84.9$_{(0.03)}$ & 87.4 \\
GLUCOSE-RoBERTa-Large & \textbf{85.7}$_{(0.03)}$ & 88.8 \\
\bottomrule
\end{tabular}
}
\caption{COPA-test Accuracy results.}
\label{tab:copa-results}
\end{table}

It is also worth mentioning three points when we compare our models with other models on COPA. First, our models, BERT-Large and RoBERTa-Large, have a significantly lower number of parameters than state-of-the-art models, Google T5-11B ($\sim$32x) and DeBERTa-1.5B ($\sim$4x) and it shows how smaller models can be competitive and benefit from continual pretraining. Second, we have not yet applied any model improvement methods such as using a margin-based loss introduced by~\citet{li2019learning} and used in CausalBERT~\cite{li2020guided}, an extra regularization loss proposed by~\citet{han-wang-2021-good}, or fine-tuning with quality-enhanced training data, BCOPA, introduced by~\citet{kavumba2019choosing}. As a result, there is still great room to improve current models that can be a proper next step. Third, we achieved performance on par with BERT-SocialIQA~\cite{sap-etal-2019-social}~\footnote{Best random seed runs on BERT and RoBERTa models achieved \%81.8 and \%88.8 accuracies, respectively.} while we did not use crowdsourcing or any \textit{manual} re-writing/correction, which is expensive, for verbalizing KG triples to create our pretraining data.

We also evaluated the performance of our models on the \textit{Easy} and \textit{Hard} question splits in COPA-test separated by~\citet{kavumba2019choosing} to see how our models perform on harder questions that do not contain superficial cues. Results are shown in Table~\ref{tab:easy-hard-results}. As can be seen, our models significantly outperformed our baselines not only on Easy questions but Hard questions as well.

\begin{table}[h]
\centering
\scalebox{0.74}{
\begin{tabular}{lcc}
\toprule
\multicolumn{1}{c}{\textbf{Model}} & \textbf{Easy} & \textbf{Hard} \\
\midrule
%\cite{han-wang-2021-good} & - & 69.7 \\
BERT-Large~\cite{kavumba2019choosing} & 83.9$_{(0.04)}$ & 71.9$_{(0.03)}$ \\
RoBERTa-Large~\cite{kavumba2019choosing} & 91.6$_{(0.01)}$ & 85.3$_{(0.02)}$ \\\midrule\midrule
\textbf{Our Models} && \\
BERT-Large (baseline) & 84.7$_{(0.05)}$ & 69.8$_{(0.09)}$ \\
ATOMIC-BERT-Large &  & \\
\hspace{10mm}\small{{- Event, Physical, Social}} & 90.6$_{(0.02)}$ & 72.1$_{(0.03)}$ \\
\hspace{10mm}\small{{- Event only}} & 88.6$_{(0.02)}$ & 73.2$_{(0.02)}$ \\
GLUCOSE-BERT-Large & 89.1$_{(0.02)}$ & 74.2$_{(0.03)}$ \\ \midrule
RoBERTa-Large (baseline) & 80.5$_{(0.01)}$ & 70.2$_{(0.12)}$ \\
ATOMIC-RoBERTa-Large &  \\
\hspace{10mm}\small{{- Event, Physical, Social}} & 87.5$_{(0.02)}$ & 81.7$_{(0.03)}$ \\
\hspace{10mm}\small{{- Event only}} & \textbf{90.7}$_{(0.03)}$ & 81.3$_{(0.04)}$ \\
GLUCOSE-RoBERTa-Large & 89.6$_{(0.05)}$ & \textbf{83.3}$_{(0.03)}$ \\
\bottomrule
\end{tabular}
}
\caption{COPA-test Accuracy results on Easy and Hard question subsets.}
\label{tab:easy-hard-results}
\end{table}

\begin{table}[h]
\centering
\scalebox{0.9}{
\begin{tabular}{lc}
\toprule
\multicolumn{1}{c}{\textbf{Model}} & \textbf{Acc (\%)} \\ \hline
%BERT-large~\cite{han-wang-2021-good} & 51.5 \\
b-l-\textit{aug}~\cite{han-wang-2021-good} & 51.1 \\
b-l-\textit{reg}~\cite{han-wang-2021-good} & 64.1 \\ \midrule \midrule
\textbf{Our Models} & \\
BERT-Large (baseline) & 51.5$_{(0.01)}$ \\
ATOMIC-BERT-Large &  \\
\hspace{10mm}\small{{- Event only}} & 53.2$_{(0.01)}$ \\
\hspace{10mm}\small{{- Event, Physical, Social}} & 53.5$_{(0.02)}$ \\
GLUCOSE-BERT-Large & \textbf{54.7}$_{(0.02)}$ \\\midrule
RoBERTa-Large (baseline) & 56.5$_{(0.06)}$ \\
ATOMIC-RoBERTa-Large &  \\
\hspace{10mm}\small{{- Event only}} & 64.2$_{(0.04)}$ \\
\hspace{10mm}\small{{- Event, Physical, Social}} & 61.8$_{(0.04)}$ \\
GLUCOSE-RoBERTa-Large & \textbf{66.1}$_{(0.03)}$ \\
\bottomrule
\end{tabular}
}
\caption{BCOPA-CE Accuracy results. Base model in \textit{b-l-*} is BERT-Large.}
\label{tab:bcopa-results}
\end{table}

\subsection{BCOPA-CE: Prompt vs. No Prompt}
\label{sec:prompt}
Results of experiments on BCOPA-CE are shown in Table~\ref{tab:bcopa-results}. As expected based on the results also reported by~\citet{han-wang-2021-good}, we initially observed that our models are performing nearly as random baseline. Since we do not use the type of question when encoding input sequences, we decided to see whether adding question type as a prompt to input sequences will improve the performance. We added {\tt It is because} and {\tt As a result,} as prompt for {\tt asks-for="cause"} and {\tt asks-for="effect"}, respectively. We observed that the new models outperformed the baseline, and our best performing model achieved a better performance than \citet{han-wang-2021-good}'s \textit{b-l-aug} and \textit{b-l-reg} models --that are fine-tuned with the same data as ours-- when question types are added as prompts to input sequences of correct and incorrect answers in the test set.
%We also ran a similar experiment on COPA-test (Table~\ref{tab:prompt-results}) in which adding a prompt did not help with performance improvement.

%\begin{table}[h]
%\centering
%\scalebox{0.9}{
%\begin{tabular}{ccc}  
%\toprule
%Train / Test & \ding{55} Prompt & \ding{51} Prompt \\
%\midrule
%\ding{55} Prompt & \textbf{79.2} & 76.4 \\
%\ding{51} Prompt & 75.5 & 77.9 \\
%\bottomrule
%\end{tabular}
%}
%\caption{COPA-test Accuracy ablation study results for prompt vs. no prompt.}
%\label{tab:prompt-results}
%\end{table}

\section{Conclusion}
\label{sec:conclusion}
We introduced a simple framework for augmenting PLMs with commonsense knowledge created by automatically verbalizing ATOMIC$^{20}_{20}$ and GLUCOSE. Our results show that commonsense knowledge-augmented PLMs outperform the original PLMs on cause-effect pair classification and answering commonsense causal reasoning questions. As the next step, it would be interesting to see how the previously proposed model improvement methods or using unbiased fine-tuning datasets can potentially enhance the performance of our knowledge-augmented models.

%\section*{Acknowledgements}

%This document has been adapted
%by Steven Bethard, Ryan Cotterell and Rui Yan
%from the instructions for earlier ACL and NAACL proceedings, including those for 
%ACL 2019 by Douwe Kiela and Ivan Vuli\'{c},
%NAACL 2019 by Stephanie Lukin and Alla Roskovskaya, 
%ACL 2018 by Shay Cohen, Kevin Gimpel, and Wei Lu, 
%NAACL 2018 by Margaret Mitchell and Stephanie Lukin,
%Bib\TeX{} suggestions for (NA)ACL 2017/2018 from Jason Eisner,
%ACL 2017 by Dan Gildea and Min-Yen Kan, 
%NAACL 2017 by Margaret Mitchell, 
%ACL 2012 by Maggie Li and Michael White, 
%ACL 2010 by Jing-Shin Chang and Philipp Koehn, 
%ACL 2008 by Johanna D. Moore, Simone Teufel, James Allan, and Sadaoki Furui, 
%ACL 2005 by Hwee Tou Ng and Kemal Oflazer, 
%ACL 2002 by Eugene Charniak and Dekang Lin, 
%and earlier ACL and EACL formats written by several people, including
%John Chen, Henry S. Thompson and Donald Walker.
%Additional elements were taken from the formatting instructions of the \emph{International Joint Conference on Artificial Intelligence} and the \emph{Conference on Computer Vision and Pattern Recognition}.

\bibliography{acl}
\bibliographystyle{acl}

% \iffalse
\appendix
\section{Contribution of Augmented Knowledge}

\begin{table*}[t!]
\centering
\scalebox{0.8}{
\begin{tabularx}{\textwidth}{X|X}
\toprule
\multicolumn{1}{c}{\textbf{COPA Test Sample}} & \multicolumn{1}{c}{\textbf{GLUCOSE Similar Entry}} \\ \hline
The family went to~\colorbox{Gray}{the zoo}. The \colorbox{Gray}{children admired the animals}. \textbf{(ask-for=result)} & The \colorbox{Green}{kids are excited} to see they are \colorbox{Green}{at the zoo} because the \colorbox{Green}{kids like(s) the zoo.} \\ \hline
The \colorbox{Gray}{phone rang}. The man \colorbox{Gray}{picked up the phone}. \textbf{(ask-for=result)} & The guy \colorbox{Green}{answers the phone} because the \colorbox{Green}{phone is ringing.} \\ \hline
The trash \colorbox{Gray}{bag was full}. I \colorbox{Gray}{took it} to the dumpster.  \textbf{(ask-for=result)} & I \colorbox{Green}{pick up the bag} since the \colorbox{Green}{trash bag is full.} \\ \hline
The runner sensed \colorbox{Gray}{his competitor gaining on} him. He \colorbox{Gray}{sped up his pace.} \textbf{(ask-for=result)} & Sam \colorbox{Green}{ran as fast as} he could since sam \colorbox{Green}{feel(s) competitive.} \\ \hline
The man \colorbox{Gray}{got out of the shower.} The \colorbox{Gray}{hot water was gone.} \textbf{(ask-for=cause)} &
All the \colorbox{Green}{hot water is gone} because my wife \colorbox{Green}{just used the shower.} \\ \hline
The \colorbox{Gray}{criminal was executed}. He was \colorbox{Gray}{convicted of murder.} \textbf{(ask-for=cause)} & The judge \colorbox{Green}{convicts} him because he is \colorbox{Green}{guilty.} \\ \hline
The boy's \colorbox{Gray}{forehead felt hot.} His \colorbox{Gray}{mother took his temperature.} \textbf{(ask-for=result)} & \colorbox{Green}{Sean's mom takes his temperature} caused sean's mom finds out \colorbox{Green}{he has a fever.} \\ \hline
The \colorbox{Gray}{fish bit the line.} The \colorbox{Gray}{fisherman reeled in the fish.} \textbf{(ask-for=result)} &
A huge \colorbox{Green}{fish gets on the line.} As a result \colorbox{Green}{bob has a bite.} \\ \hline
The man \colorbox{Gray}{went to the doctor.} The man \colorbox{Gray}{felt ill.} \textbf{(ask-for=cause)} &
Tom \colorbox{Green}{goes to the doctor} because tom \colorbox{Green}{feel(s) sick.} \\ \hline
An \colorbox{Gray}{unfamiliar car} parked outside my house. I \colorbox{Gray}{became suspicious.} \textbf{(ask-for=result)} &
I notice an \colorbox{Green}{unfamiliar car.} As a result I \colorbox{Green}{feel(s) curiosity.} \\
\bottomrule
\end{tabularx}
}
\caption{Correctly classified samples in COPA and their most semantically similar entries in GLUCOSE.}
\label{tab:copa-error-analysis}
\end{table*}

We did further analysis to better understand how the augmented knowledge did or did not help PLMs in achieving better results on our benchmarks. Even though knowing how exactly data points from ATOMIC$^{20}_{20}$ and GLUCOSE contributed to performance improvements is hard and may need a more rigorous analysis, we found it helpful to investigate the semantic overlap between the augmented data and our benchmarks' samples to see if the injected knowledge has any context similarity with what our models were tested on. In each benchmark, we picked our best performing model and the baseline and separated all samples in the test set that were correctly predicted across \textit{all} random seed runs by these models. Then, we created a set of correctly predicted samples by our best model that our baseline failed to predict correctly. And we measured the semantic similarity of each sample in that set with all data points in ATOMIC$^{20}_{20}$ and GLUCOSE. To measure semantic similarity, we leveraged the {\tt Sentence Transformers}~\cite{reimers-2019-sentence-bert}.\footnote{\url{https://github.com/UKPLab/sentence-transformers}} In particular, after computing the embeddings of samples,\footnote{The model we use is available on HuggingFace: {\tt sentence-transformers/all-mpnet-base-v2}} we computed the cosine similarity between pairs of embeddings and separated pairs with at least \%50 similarity. Our idea was that if we had a data point in ATOMIC$^{20}_{20}$ or GLUCOSE that has a high semantic similarly ---in terms of the interactions between events--- with a data point in the benchmark, that semantic similarity may have contributed to the augmented model's performance improvement.

Table~\ref{tab:copa-error-analysis} shows examples of the correctly classified samples with high context similarity with entries in GLUCOSE. Out of 70,730 training samples in GLUCOSE, there are 3,588 and 253 pairs with 0.5 and 0.6 cosine similarity with a sample in COPA, respectively. As can be seen, there is not necessarily an exact match but a context similarity between samples in each pair. For instance, from an entry in GLUCOSE we know that \textit{noticing an unfamiliar car} will result in \textit{feeling curios}. And this is what has been asked in a question in COPA where \textit{being suspicious} is the plausible result of seeing \textit{an unfamiliar car parked outside house}. Such examples suggest that a model may have learned the relation between \textit{seeing an unfamiliar object} and \textit{a curiosity feeling} at the time of continual pretraining which helped it later to predict the correct answer when two similar events are involved in a question. It is worth emphasizing that we may not be able to claim that this context similarity is the cause for the performance enhancement of augmented models, however, it is still interesting to see that feeding a model with explicit causal statements potentially helps the model to express the causal knowledge that may or may not already be encoded in the model, as also stated in previous work~\cite{Hwang2021COMETATOMIC2O}.

%\section{Hyperparameters}
%For reproducibility purposes, we list the hyperparameters we used for pretraining and fine-tuning in Table~\ref{tab:pretraining-hyperparameters} and Table~\ref{tab:finetuning-hyperparameters}, respectively.

%\begin{table}[h]
%\centering
%\begin{tabular}{lccc}
%\toprule
%\textbf{Model}  & \textbf{lr}& \textbf{epoch}& %\textbf{batch} \\ \hline
%BERT-Large-G & 2e-5 & 25 & 32 \\
%\bottomrule
%\end{tabular}
%\caption{Pretraining Hyperparameters}
%\label{tab:pretraining-hyperparameters}
%\end{table}

%\begin{table}[h]
%\centering
%\begin{tabular}{lccc}
%\toprule
%\textbf{Model}  & \textbf{lr}& \textbf{epoch}& \textbf{batch} \\ \hline
%BERT-Large & 2e-5 & 4 & 4 \\
%GLUCOSE-BERT-Large & 2e-5 & 4 & 4 \\
%ATOMIC-BERT-Large & 2e-5 & 4 & 4 \\
%\bottomrule
%\end{tabular}
%\caption{Fine-tuning best hyperparameters}
%\label{tab:finetuning-hyperparameters}
%\end{table}
% \fi
%\section{Appendix}
%\label{sec:appendix}

%This is an appendix.

%\section*{Acknowledgements}

\end{document}